\documentclass[10pt,twocolumn,letterpaper]{article}

\usepackage{colortbl}
\usepackage{multirow}
\usepackage{utfsym}

\usepackage[pagenumbers]{cvpr} 

\definecolor{cvprblue}{rgb}{0.21,0.49,0.74}
\usepackage[pagebackref,breaklinks,colorlinks,allcolors=cvprblue]{hyperref}

\def\paperID{4237} 
\def\confName{CVPR}
\def\confYear{2025}

\title{CLIPer: Hierarchically Improving Spatial Representation of CLIP for Open-Vocabulary Semantic Segmentation}

\author{Lin Sun$^{1}$, Jiale Cao$^{1}$, Jin Xie$^{2}$, Xiaoheng Jiang$^{3}$, Yanwei Pang$^{1, 4}$\\
    $^1$Tianjin University~~~$^2$Chongqing University~~~$^3$Zhengzhou University\\
    $^4$Shanghai Artificial Intelligence Laboratory\\
    {\tt\small \{sun0806,~connor,~pyw\}@tju.edu.cn} \\
    {\tt\small xiejin@cqu.edu.cn,~jiangxiaoheng@zzu.edu.cn}
}

\begin{document}
\maketitle
\begin{abstract}
Contrastive Language-Image Pre-training (CLIP) exhibits strong zero-shot classification ability on various image-level tasks, leading to the research to adapt CLIP for pixel-level open-vocabulary semantic segmentation without additional training. The key is  to improve spatial representation of image-level CLIP, such as replacing self-attention map at last layer with self-self attention map or vision foundation model based attention map. In this paper, we present a novel hierarchical framework, named CLIPer, that hierarchically improves spatial representation of CLIP. The proposed CLIPer includes an early-layer fusion module and a fine-grained compensation module. We observe that, the embeddings and attention maps at early layers can preserve spatial structural information. Inspired by this, we design the early-layer fusion module to generate  segmentation map with better spatial coherence. Afterwards, we employ a fine-grained compensation module to compensate the local details using the self-attention maps of diffusion model. We conduct the experiments on seven segmentation datasets. Our proposed  CLIPer achieves the state-of-the-art performance on these datasets. For instance, using ViT-L, CLIPer has the mIoU  of 69.8\% and 43.3\% on VOC and COCO Object, outperforming ProxyCLIP by 9.2\% and 4.1\% respectively. We release the source code and models at \url{https://linsun449.github.io/cliper}.

\end{abstract}    
\section{Introduction}
\label{sec:intro}

\begin{figure}[t]
\centering
\includegraphics[width=\linewidth]{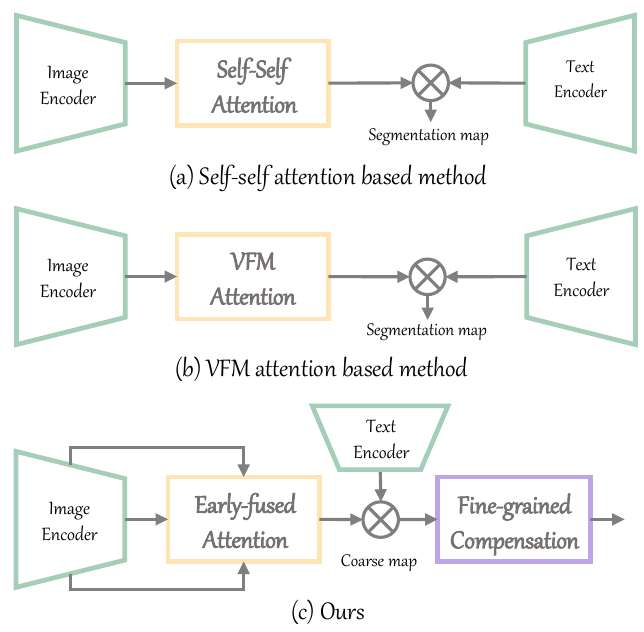}
\caption{Comparison with existing CLIP-based open-vocabulary semantic segmentation approaches without training. In (a), several approaches \cite{Wang_2024_SCLIP,Lan_2024_ClearCLIP,maskclip} replace the original self-attention map at last layer with self-self attention map, which can better maintain spatial coherence.  In (b), the method ProxyCLIP \cite{Lan_2014_ProxyCLIP} opts for a different strategy,  replacing original self-attention map with vision foundation model-based (VFM-based) attention map. In (c), we utilize the embeddings and  self-attention maps at early layers to fully exploit spatial information within CLIP. Subsequently, we perform fine-grained compensation using diffusion model to further improve local details.}
\vspace{-6pt}
\label{fig:intro}
\end{figure}

Open-vocabulary semantic segmentation \cite{Zhou_2022_MaskCLIP,ZS3Net,zhao2017open} aims to divide an image into different groups and assign each group a label belonging to arbitrary semantic categories. Compared to the traditional semantic segmentation, open-vocabulary semantic segmentation is a more challenging segmentation task. Recently, the researchers mainly explored to employ vision-language models for open-vocabulary semantic segmentation. The related methods can be divided into training-based \cite{san,fcclip,catseg} and training-free \cite{Zhou_2022_MaskCLIP,Wang_2024_SCLIP,Lan_2024_ClearCLIP} approaches. Compared to  training-based approaches,  training-free counterparts are simpler.

Contrastive Language-Image Pre-training (CLIP) model \cite{Radford_2021_CLIP} has shown strong zero-shot capabilities on image-level classification task, due to the pre-training on large-scale image-text paired data \cite{Schuhmann_2022_Laion}. Based on this, several  methods have been proposed to adapt CLIP for training-free open-vocabulary semantic segmentation. The key challenge is to improve spatial representation of image-level supervised model for pixel-level segmentation. As shown in Fig. \ref{fig:intro}(a), some methods modify the original self-attention map at last layer with self-self attention map, better maintaining local spatial information. For instance, MaskCLIP \cite{Zhou_2022_MaskCLIP} employs an identical
self-self matrix as the self-attention map at last layer to generate visual patch embeddings, while SCLIP \cite{Wang_2024_SCLIP} and  ClearCLIP \cite{Lan_2024_ClearCLIP} employ the  query-to-query or key-to-key attention map to replace original self-attention map. Instead of using original or self-self attention map, ProxyCLIP \cite{Lan_2014_ProxyCLIP} extracts the self-attention map from visual foundation model (VFM) \cite{Caron_2021_DINO} as self-attention map at last layer in Fig. \ref{fig:intro}(b). These methods  enhance segmentation performance of CLIP in open-vocabulary setting without additional training. However, these methods mainly consider improving the self-attention map at last layer of CLIP.

In this paper, we focus on two factors to hierarchically  improve spatial representation. (i) The first one is to improve patch-level spatial coherence similar to existing methods \cite{maskclip,Lan_2024_ClearCLIP}. We observe that, the patch embeddings and attention maps  at early layers contain rich spatial structural information. Therefore, instead of using self-self   or VFM-based attention at last layer, we aim to exploit early-layer information of CLIP.  
(ii) The second is fine-grained compensation. The patch-level similarity map between image and text is relatively  coarse in local details. It is necessary to further improve local details for improved segmentation.

Based on these two factors above, we introduce a novel hi\underline{er}archical method for open-vocabulary semantic segmentation, named CLIPer. Our CLIPer consists of an early-layer fusion module and a fine-grained compensation module. As shown in Fig. \ref{fig:intro}(c), the early-layer fusion module integrates patch embeddings and attention maps from early layers to improve spatial coherence of output patch embeddings. Based on the output patch embeddings and text embeddings of arbitrary categories, we can generate the corse segmentation map. Afterwards, the fine-grained compensation module integrates fine spatial information of Stable Diffusion to compensate the local details. We conduct experiments on various segmentation datasets. The contributions and merits are summarized as

\begin{itemize}
    \item We propose a novel training-free CLIP-based method that hierarchically improves the spatial representation of CLIP for open-vocabulary sematic segmentation.
    \item An early-layer fusion strategy is introduced to improve patch-level coherence within CLIP by integrating early-layer information.
    \item A fine-grained compensation module leverages the fine detail information from diffusion model to  refine local details lost in CLIP, leading to more precise segmentation.
    \item Our  method achieves the superior performance on various segmentation datasets. For instance,  using ViT-L backbone, it achieves mIoU scores of 69.8\% and 43.3\% on VOC and Object, respectively.
\end{itemize}

\section{Related Work}
\label{sec:relatedwork}

\noindent\textbf{Vision-language pre-training models.}
Vision-language pre-training models aim to establish relationships between the images and texts. Among these models, CLIP \cite{Radford_2021_CLIP} is one of most successful vision-language models, which is trained on a very large-scale image-text paired dataset. Due to the large-scale pre-training, CLIP exhibits strong zero-shot classification performance on various image-level tasks. OpenCLIP \cite{cherti2023OpenCLIP} explores to improve CLIP via conducting a comprehensive  experimental analysis of scaling laws. To address the challenge of expensive image-text annotations, ALIGN \cite{align} introduces to use large-scale noisy image-text data for model learning.

\noindent\textbf{Open-vocabulary semantic segmentation.} Compared to traditional semantic segmentation \cite{Xie_2021_SegFormer,Long_2015_FCN} sharing a fixed category set between the training and test sets, open-vocabulary semantic segmentation \cite{Shin_2022_RECO,Lin_2023_CLIPES,Kanchana_2022_CLIPpy} aims to segment the objects belonging to arbitrary  categories. In the past years, open-vocabulary semantic segmentation has attracted great attention, and achieved substantial progress. The related methods can be divided into training-based \cite{Luo_2023_SegCLIP,Ren_2023_VIEWCO,Xu_2023_OVSegmenter,cha2022tcl} and training-free \cite{Wang_2024_SCLIP,Lin_2024_TagCLIP} approaches. Training-based methods first train a model on a fixed set of categories from a given training dataset and then apply the learned model to segment objects from arbitrary semantic categories. Some training-based approaches \cite{ovseg,odise} follow two-stage pipeline, where the first stage extracts the mask proposals, and the second stage assigns  semantic labels to mask proposals. For instance, OVSeg \cite{ovseg} first trains a class-agnostic mask proposals using query-based framework Mask2Former \cite{cheng2022masked}, and then fine-tunes CLIP to classify the cropped and masked images. Some training-based approaches adopt single-stage pipeline. For instance, SAN \cite{san} introduces a side adapter to adapt CLIP for both classification and segmentation. SED \cite{Xie_2024_SED} introduces a simple encoder-decoder architecture  with category early rejection for fast inference. CAT-Seg \cite{catseg} constructs pixel-level cost map for segmentation. SAM-CLIP \cite{Wang_2024_SAMCLIP} integrates CLIP and SAM \cite{kirillov2023segment} into a single multi-task segmentation.  

Compared to training-based approaches, training-free methods aim to directly adapt vision-language models for open-vocabulary semantic segmentation without any training. Most training-free approaches focus on exploring to improve spatial coherence of image-level supervised CLIP. For instance, MaskCLIP \cite{Zhou_2022_MaskCLIP} removes the self-attention at last layer, and directly employ value embeddings as output embeddings to perform pixel-level segmentation. Instead of removing self-attention, SCLIP \cite{Wang_2024_SCLIP} and ClearCLIP \cite{Lan_2024_ClearCLIP} employ query-to-query or key-to-key attention map to replace original attention map at last layer. ProxyCLIP \cite{Lan_2014_ProxyCLIP} first calculates the self-attention map of vision foundation model. CaR \cite{Sun2024CaR} adopts a recurrent framework to progressively  enhance segmentation.

In addition, some researchers have  explored to employ diffusion models for open-vocabulary semantic segmentation. For instance, ODISE \cite{odise} employs diffusion model to generate mask proposals, and generates the visual embeddings of masks for classification. OVDiff \cite{Karazija_2023_OVDiff} generates support images of arbitrary categories using diffusion model, and extract the features of prototypes  to segment inference images. DiffSegmenter \cite{Wang_2023_DiffSegmenter} and iSeg \cite{Sun_2024_iSeg} exploit self-attention and cross-attention maps from diffusion models for open-vocabulary segmentation. 

In this paper, we explore to hierarchically improve spatial representation of CLIP for open-vocabulary semantic segmentation.  CLIP demonstrates  better zero-shot classification performance, while diffusion model is effective in capturing local details. Based on this, we first adopt CLIP to extract coarse segmentation map, and second employ diffusion model to refine the local details of segmentation map.

\section{Methodology}\label{sec:method}

Here, we first give some preliminaries of CLIP, Stable Diffusion, and attention mechanism. Afterwards, we introduce the motivation and our proposed method.

\subsection{Preliminary}
\label{preliminary}

\textbf{CLIP.} CLIP \cite{Radford_2021_CLIP} contains an image encoder and a text encoder. The image encoder comprises of a series of transformer blocks \cite{vit}, where the input image is divided into patches and processed through these blocks. Each transformer block consists of a residual attention and a residual FFN. Initially, a class token is added to aggregate information from all image patches, forming a global representation of the image. Subsequently, each transformer block processes an input embeddings $F = \left [F_{cls}, F_1, ..., F_{h\times w}  \right ]$, where $F_{cls}$ represents the embeddings of class token, and others correspond to the embeddings of patch tokens. The output embeddings of class token is finally aligned with the embeddings generated by text encoder.

\noindent\textbf{Stable Diffusion.} Image diffusion model generates images starting from random Gaussian noise  through a series of denoising steps. By training on large-scale dataset, the diffusion model Stable Diffusion \cite{Rombach_2021_LDM} is able to generate high-quality images with rich  details. It has been shown that,  the features in Stable Diffusion are able to  accurately capture local detail information. Therefore, we explore to use Stable Diffusion to improve local details of segmentation.

\noindent\textbf{Attention mechanism.} Both CLIP and Stable Diffusion leverage attention mechanism. Specifically, CLIP employs the self-attention to model relationships between image patches. In contrast, Stable Diffusion incorporates both self-attention and cross-attention, where self-attention is used to extract spatial coherence within the image, and cross-attention allows the model to incorporate external conditioning information (\textit{e.g.,} text description) to guide the image generation.
The output of attention mechanism, whether in self-attention or cross-attention, is calculated by the query $Q$, key $K$ and value $V$ as follows
\begin{equation}
\mathrm{Att}(Q, K, V) = A \times V,
\end{equation}
the attention map $A$ is given by
\begin{equation}
A = \mathrm{Softmax}(\frac{QK^{\mathrm{T}}}{\sqrt{d}}),
\label{eq:attn_map}
\end{equation}
where $d$ is the feature dimensionality of the key $K$.
In self-attention, the query, key, and value all come from the same embeddings of image. In contrast, in cross-attention, the query comes from image, while the key and value come from text description.

\subsection{Motivation}
\label{motivation}

\begin{figure}[t]
\centering
\includegraphics[width=\linewidth]{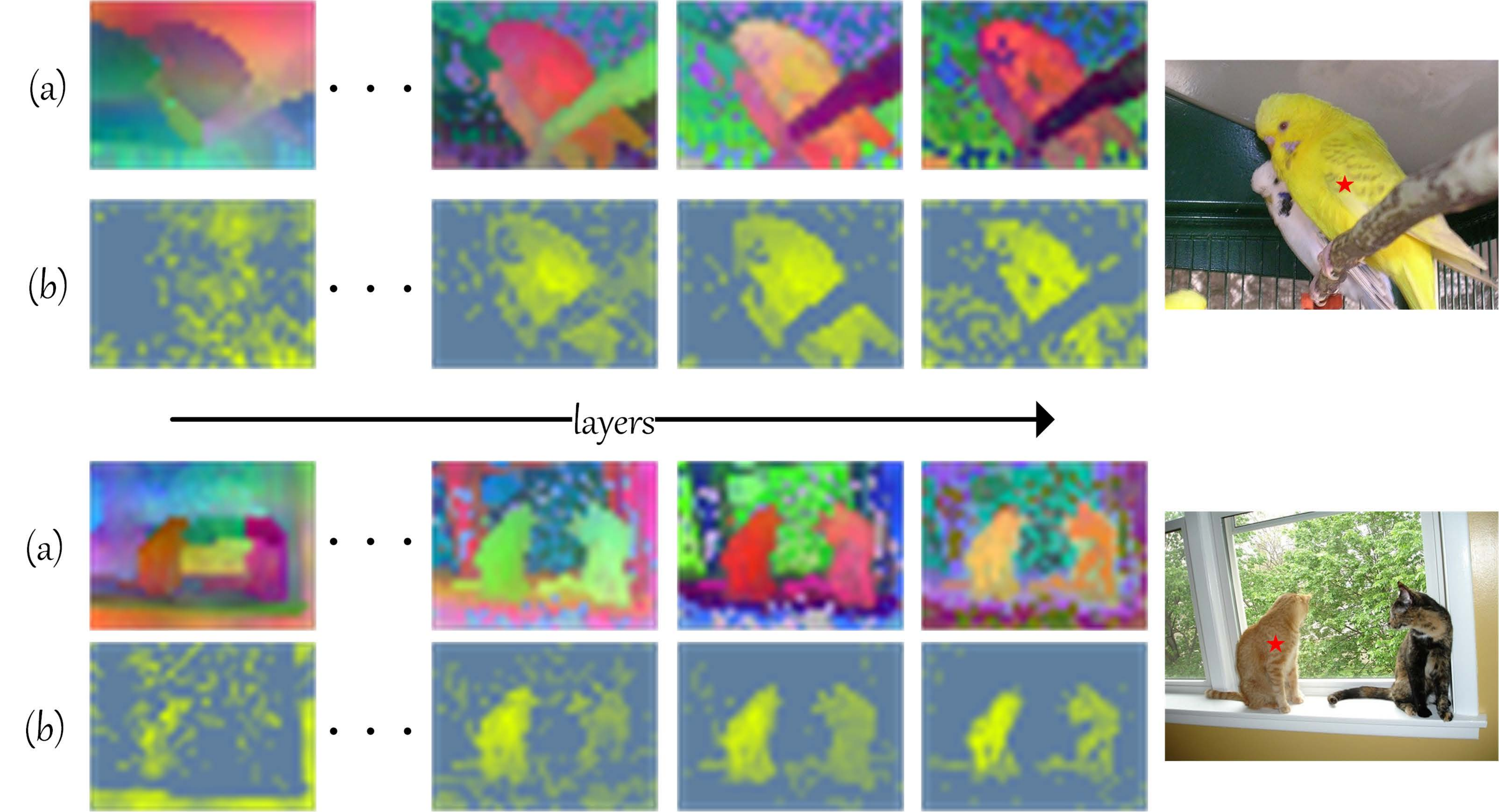}
\caption{Visualization of path embeddings in CLIP. In (a), we visualize the embeddings into a 3D space and observe that the early embeddings exhibit good spatial coherence. In (b), we evaluate the cosine similarity between the embeddings of early layers at a specific point and the embeddings at last layer, revealing that the earlier and last embeddings  share a similar embedding space.}
\label{fig:embedings}
\end{figure}

\begin{figure}[t]
\centering
\includegraphics[width=\linewidth]{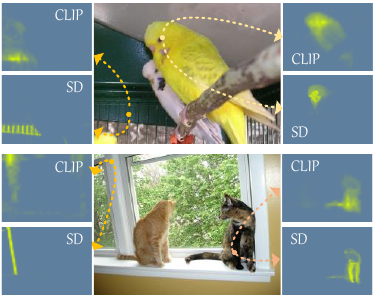}
\caption{Visualization of self-attention maps between CLIP and Stable Diffusion (SD). We show the self-attention maps at selected points for both CLIP and SD. Compared to that of CLIP, we observe that the self-attention maps of SD  focus more on capturing local details.}
\label{fig:attentions}
\vspace{-6pt}
\end{figure}

To adapt pre-trained CLIP for open-vocabulary segmentation, one straightforward approach is to discard the class token and use only the patch tokens to generate pixel-level similarity map with text embeddings. However, since CLIP is pre-trained on image-level classification task, this simple approach usually achieves poor segmentation due to weak spatial coherence of patch embeddings. To address this issue, some  approaches \cite{maskclip,Wang_2024_SCLIP} primarily focus on modifying the last layer to improve spatial coherence. In contrast, we propose to hierarchically improve spatial representation for better segmentation from two aspects of observations.

\begin{figure*}[t]
\centering
\includegraphics[width=0.8\linewidth]{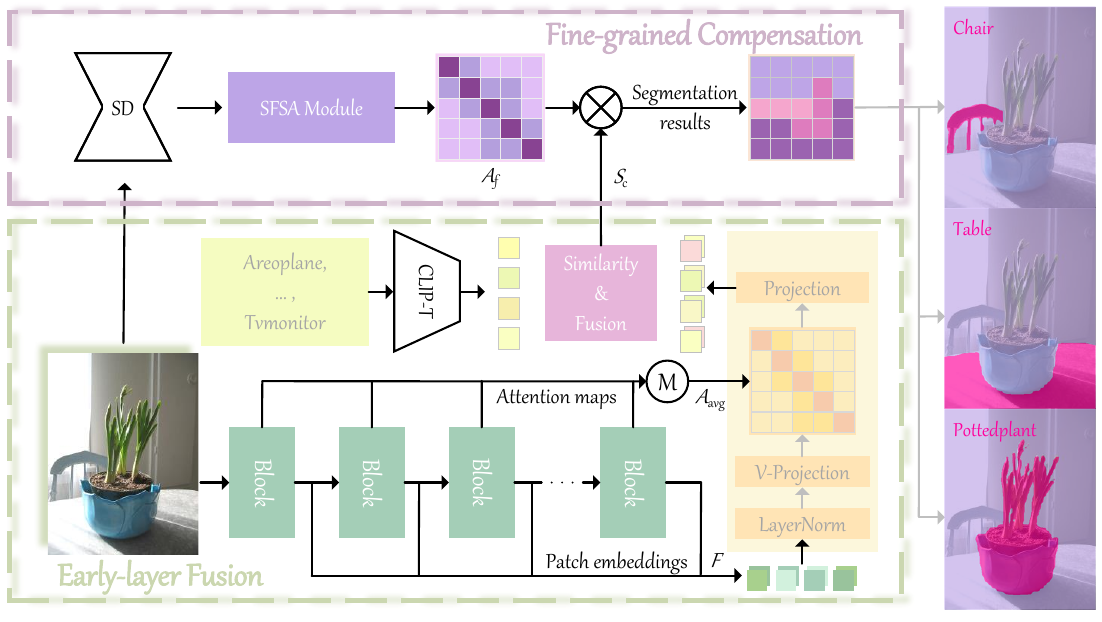}
\caption{Overall architecture of our proposed method CLIPer. Our CLIPer contains two components: early-layer fusion and fine-grained compensation. In the early-layer fusion, we  aggregate early-layer information of CLIP image encoder, including embeddings and attention maps, to improve spatial coherence of output embeddings, which are used to generate coarse segmentation map with text embeddings. The fine-grained compensation aims to employ self-attention maps of Stable Diffusion to refine local details of coarse segmentation map.}  
\label{fig:framework}
\vspace{-6pt}
\end{figure*}

We observe that the embeddings at early layers are suitable for improving spatial coherence. First, as shown in Fig. \ref{fig:embedings}(a), the patch embeddings at early layers  retain  consistent spatial information. Second, as in Fig. \ref{fig:embedings}(b), early-layer embeddings share similarities with the embeddings at last layer.  This similarity enables effective fusion of early- and late-layer information.

The pre-trained Stable Diffusion can generate high-quality images with rich details. As  in Fig. \ref{fig:attentions}, by visualizing the  self-attention maps in both CLIP and Stable Diffusion, we observe that the attention maps of Stable Diffusion effectively capture local details. This contrasts with the attention maps in CLIP, which typically respond to broader semantic areas. This key observation suggests that we can integrate the fine spatial information from Stable Diffusion to improve coarse segmentation generated by CLIP.

\subsection{Framework}
\label{framework}
\textbf{Overview.} Inspired by the motivation above, we propose a novel hierarchical approach for open-vocabulary semantic segmentation. Fig. \ref{fig:framework} presents an overall architecture of our proposed method, named CLIPer. Our CLIPer consists of two complementary components. In  first component, we leverage an early-layer fusion module to generate patch embeddings with better spatial coherence, and then generate coarse segmentation map according to the similarity between patch embeddings and text embeddings of text encoder. In second component, we perform fine-grained compensation using the attention maps of Stable Diffusion.

\noindent\textbf{Early-layer fusion.} This module aims to improve spatail coherence using the embeddings of early layers. Specifically, given an image, we first divide the image  into patch embeddings $F^0 \in \mathbb{R}^{(hw + 1) \times D}$, and then feed these embeddings to a series of transformer blocks. For the $n$-th transformer block, we generate the query $Q^n$, key $K^n$, and value $V^n$ as
\begin{equation}
Q^n = \mathrm{Proj_q}(\mathrm{LN}(F^{n-1})),
\end{equation}
\begin{equation}
K^n = \mathrm{Proj_k}(\mathrm{LN}(F^{n-1})),
\end{equation}
\begin{equation}
V^n = \mathrm{Proj_v}(\mathrm{LN}(F^{n-1})).
\end{equation}

Here, LN denotes layer normalization, and the projections $\mathrm{Proj_q}$, $\mathrm{Proj_k}$, and $\mathrm{Proj_v}$ respectively contain one learnable linear layer. With  the query $Q^n$, key $K^n$, and value $V^n$, the  output embeddings $F^n$ of $n$-th transformer block are calculated by
\begin{equation}
\Bar{F}^n = \mathrm{Att}(Q^n, K^n, V^n) + F^{n-1},
\label{eq:ffn}
\end{equation}
\begin{equation}
F^n = \mathrm{FFN}(\mathrm{LN}(\Bar{F}^n)) + \Bar{F}^n,
\end{equation}
where FFN stands for a feed-forward network.
The attention map  of $n$-th transformer, which is generated  in attention operation of Eq. \ref{eq:ffn}, are denoted as $A^n$. 

Similarly, we can generate the embeddings and attention maps of all transformer blocks up to the penultimate layer, denoted as two sets: $\mathcal{F} = \left \{F^i | i=1, 2, ..., N-1 \right \}$ and $\mathcal{A} = \left \{A^i | i=1, 2, ..., N-1 \right \}$. We first generate an averaged attention map  as
\begin{equation}
A_{avg} = \frac{1}{N} \sum_{n=1}^{N-1} A^n.
\end{equation}
Then, we replace the original self-attention map at last layer with the averaged attention map $A_{avg}$, and then feed all the embeddings to the last layer. Similar to ClearCLIP \cite{Lan_2024_ClearCLIP}, we omit the feed-forward network and residual connections in the last transformer block, which can simplify the representation while aligning text embeddings better. As a result, we generate multiple output embeddings for different layers. 

Lastly, we compute the cosine similarity between multiple output embeddings and text embeddings derived from the CLIP text encoder, and calculate the averaged similarity map. This averaged similarity map is used as the coarse segmentation by mapping each patch embedding to the candidate category embeddings.

\noindent\textbf{Fine-grained compensation.} The patch-level segmentation generated by CLIP remains relatively coarse, limiting segmentation accuracy. To better compensate local details of  coarse map, we leverage the self-attention maps from Stable Diffusion, which we find to be particularly effective at capturing fine-grained local information. This locality-preserving characteristic is highly beneficial for refining the spatial details of patch-level segmentation, improving the ability to distinguish boundaries.

Specifically, we first feed the image along with an empty (null) textual prompt into Stable Diffusion, obtaining the corresponding multi-head self-attention maps at the highest spatial resolutions. We denote this attention maps as $A_{m} \in \mathbb{R}^{H \times L \times L}$, where $H$ represents the number of attention heads, and $L$ indicates the spatial size of  feature maps in Stable Diffusion. Then, we fuse these attention maps $A_{m}$  by matrix chain multiplication across the attention heads, which is formulated as
\begin{equation}
A_{f} = A_{m}[0] \times A_{m}[1] \times \cdots \times A_{m}[H-1],
\label{eq:fused_attn}
\end{equation}
where $A_{m}[i]$ means the $i$-th head self-attention map. Afterwards, we utilize the fused attention map $A_{f}$ to refine the upscaled coarse segmentation map $S_{c}$ as
\begin{equation}
S_{f} = A_{f} \times S_{c}.
\label{eq:refine}
\end{equation}
Finally, we upscale $S_{f}$ to the resolution of input image, yielding the fine-grained pixel-level segmentation map.

\section{Experiment}
Here we conduct experiments to demonstrate the effectiveness of our proposed method on various datasets.

\subsection{Experimental Setups}
\textbf{Datasets.} We evaluate CLIPer on seven datasets similar to most existing methods: (1) Considering the background category. We use PASCAL VOC  (VOC) \cite{Everingham_2010_PASCAL}, PASCAL Context  (Context) \cite{Mottaghi_2014_Context}, and COCO Object (Object) \cite{Caesar_2018_Object_Stuff}; (2) Without considering the background category. We use PASCAL VOC (VOC20), PASCAL Context (Context59), COCO-Stuff (Stuff) \cite{Caesar_2018_Object_Stuff}, and ADE20K (ADE) \cite{ade20k}. For performance evaluation, we use the validation set from each dataset. In addition, for weakly supervised semantic segmentation, we evaluate the pseudo-mask generation performance on  the training sets of VOC  and COCO datasets.

\noindent\textbf{Metrics.} We use mean Intersection over Union (mIoU) to evaluate pixel-level segmentation accuracy. Further, we adopt mAP, F1 score, Precision (P), and Recall (R) to evaluate image-level classification performance.

\noindent\textbf{Implementation details.} 

We implement our method on a single RTX 3090 with 24G memory. We employ ViT-B and ViT-L as the backbones, and uses Stable Diffusion V2.1 \cite{Rombach_2021_LDM} for fine-grained compensation. In the Stable Diffusion, we extract the attention maps at time-step  45 in total of 50 steps.  We set text prompts including  category descriptions similar to SCLIP \cite{Wang_2024_SCLIP} and ProxyCLIP \cite{Lan_2014_ProxyCLIP}. We resize all the input images to a shorter side of 336 pixels while maintaining the original aspect ratio, similar to ProxyCLIP \cite{Lan_2014_ProxyCLIP}. Instead of using a sliding window strategy in \cite{Zhou_2022_MaskCLIP}, \cite{Wang_2024_SCLIP}, \cite{Lan_2024_ClearCLIP} and \cite{Lan_2014_ProxyCLIP}, we directly feed the entire image into the CLIP image encoder, which is faster and simplifies the process.

\begin{table*}[t]
\centering
\footnotesize
\setlength{\tabcolsep}{3.0mm}
\begin{tabular}{c|c|c|ccc|cccc}
\toprule 
Type        & Method         & Encoder    & VOC      & Context       & Object          & VOC20                    & Context59         & Stuff       & ADE          \\ \midrule
& SegCLIP \cite{Luo_2023_SegCLIP}                  & ViT-B/16                  & 52.6          & 24.7          & 26.5          & -             & -             & -             & -             \\
& ViewCo   \cite{Ren_2023_VIEWCO}                 & ViT-S/16                  & 52.4          & 23.0          & 23.5          & -             & -             & -             & -             \\
& OVSegmentor \cite{Xu_2023_OVSegmenter}             & ViT-B/16                  & 53.8          & 20.4          & 25.1          & -             & -             & -             & -             \\
& CoCu \cite{xing2023cocu}                     & ViT-S/16                  & 51.4          & 23.6          & 22.7          & -             & -             & 22.1          & 12.3          \\
& SAM-CLIP \cite{Li2023CLIPSurgery}                  & ViT-B/16                  & 60.6          & 23.2          & -             & -             & -             & -             & 17.1          \\
& GroupViT \cite{groupvit}                  & ViT-S/16                  & 50.4          & 18.7          & 27.5          & 79.7          & 23.4          & 15.3          & 9.2           \\
\multirow{-6}{*}{Training-based}& TCL     \cite{cha2022tcl}                  & ViT-B/16                  & 51.2          & 24.3          & 30.4          & 77.5          & 30.3          & 19.6          & 14.9          \\
\multirow{-6}{*}{(weakly-supervised)}    & CLIP-DINOiser    \cite{wysoczanska2024clipdino}          & ViT-B/16                  & 62.2          & 32.4          & 35.0          & 80.2          & 35.9          & 24.6          & 20.0          \\ 
\midrule
& CLIP \cite{Radford_2021_CLIP}                     & ViT-B/16                  & 16.4          & 8.4           & 5.6           & 41.9          & 9.2           & 4.4             & 2.9           \\
& CLIPSurgery \cite{Li2023CLIPSurgery}              & ViT-B/16          & -             & 29.3          & -             & -             & -             & 21.9          & -             \\
& MaskCLIP \cite{Zhou_2022_MaskCLIP}                  & ViT-B/16                  & 38.8          & 23.6          & 20.6          & 74.9          & 26.4          & 16.4            & 9.8           \\
& SCLIP \cite{Wang_2024_SCLIP}                     & ViT-B/16                  & 59.1          & 30.4          & 30.5          & 80.4          & 34.2          & 22.4            & 16.1          \\
& ClearCLIP \cite{Lan_2024_ClearCLIP}                 & ViT-B/16                  & 51.8          & 32.6          & 33.0          & 80.9          & 35.9          & 23.9            & 16.7          \\
& ProxyCLIP \cite{Lan_2014_ProxyCLIP}                & ViT-B/16                  & 61.3          & 35.3          & 37.5          & 80.3          & 39.1          & 26.5            & 20.2          \\
& \cellcolor{gray!20}\textbf{CLIPer* (Ours)}              & \cellcolor{gray!20}ViT-B/16     & \cellcolor{gray!20}60.1         &  \cellcolor{gray!20}34.8   & \cellcolor{gray!20}36.0   &  \cellcolor{gray!20}84.0      & \cellcolor{gray!20}38.5    & \cellcolor{gray!20}25.3    & \cellcolor{gray!20}19.8    \\
& \cellcolor{gray!20}\textbf{CLIPer (Ours)}              & \cellcolor{gray!20}ViT-B/16     & \cellcolor{gray!20}\textbf{65.9}         &  \cellcolor{gray!20}\textbf{37.6}  & \cellcolor{gray!20}\textbf{39.0}   &  \cellcolor{gray!20}\textbf{85.2}      & \cellcolor{gray!20}\textbf{41.7}    & \cellcolor{gray!20}\textbf{27.5}    & \cellcolor{gray!20}\textbf{21.4}    \\ 
\cline{2-10} 
& CLIP \cite{Radford_2021_CLIP}                     & ViT-L/14                  & 8.2           & 4.1           & 2.7           & 15.6          & 4.4           & 2.4             & 1.7           \\
& MaskCLIP \cite{Zhou_2022_MaskCLIP}                 & ViT-L/14                  & 23.3          & 11.7          & 7.2           & 29.4          & 12.4          & 8.8             & 7.2           \\
& SCLIP \cite{Wang_2024_SCLIP}                    & ViT-L/14                  & 43.5          & 22.3          & 25.0          & 69.1          & 25.2          & 17.6            & 10.9          \\
& ClearCLIP \cite{Lan_2024_ClearCLIP}                & ViT-L/14                  & 46.1         & 26.7         & 30.1         & 80.0          & 29.6          & 19.9            & 15.0          \\
& CaR  \cite{Sun2024CaR}              & ViT-L/14                  & 67.6         & 30.5         & 36.6         & \textbf{91.4} & 39.5          & -            & 17.7          \\
& ProxyCLIP \cite{Lan_2014_ProxyCLIP}                & ViT-L/14                  & 60.6          & 34.5          & 39.2          & 83.2          & 37.7          & 25.6            & 22.6          \\
& ProxyCLIP \cite{Lan_2014_ProxyCLIP}            & ViT-H/14          & 65.0          & 35.4          & 38.6          & 83.3          & 39.6          & 26.8          & 24.2          \\
& \cellcolor{gray!20}\textbf{CLIPer* (Ours)}     & \cellcolor{gray!20}ViT-L/14         & \cellcolor{gray!20}61.2 & \cellcolor{gray!20}34.3 & \cellcolor{gray!20}39.6 & \cellcolor{gray!20}88.2 & \cellcolor{gray!20}39.8 & \cellcolor{gray!20}25.8   & \cellcolor{gray!20}21.8 \\
\multirow{-17}{*}{Training-free}        & \cellcolor{gray!20}\textbf{CLIPer (Ours)}     & \cellcolor{gray!20}ViT-L/14         & \cellcolor{gray!20}\textbf{69.8} & \cellcolor{gray!20}\textbf{38.0} & \cellcolor{gray!20}\textbf{43.3} & \cellcolor{gray!20}90.0 & \cellcolor{gray!20}\textbf{43.6} & \cellcolor{gray!20}\textbf{28.7}   & \cellcolor{gray!20}\textbf{24.4} \\ 
\bottomrule
\end{tabular}
\caption{Comparison with existing open-vocabulary segmentation methods. * denotes our method without fine-grained compensation. Our method  achieves state-of-the-art segmentation accuracy (mIoU) on all datasets, except VOC20 with the backbone ViT-L.}
\label{tab:sota_ovs}
\vspace{-6pt}
\end{table*}

\begin{table*}[t]
\centering
\footnotesize
\setlength{\tabcolsep}{2.9mm}
\begin{tabular}{l|c|ccll|cccl|cccl}
\toprule
                          &                           & \multicolumn{4}{c|}{VOC}                                      & \multicolumn{4}{c|}{Context}                                  & \multicolumn{4}{c}{Object}                                    \\ 
                          \cmidrule(l){3-14} 
\multirow{-2}{*}{Method} & \multirow{-2}{*}{Encoder} & mAP           & F1            & P             & R             & mAP           & F1            & P             & R             & mAP           & F1            & P             & R             \\ 
\midrule
CLIP \cite{Radford_2021_CLIP}                     & ViT-L/14                  & 90.2          & 75.3          & 79.7          & 71.3          & 59.8          & 54.2          & 55.1          & 53.2          & 66.3          & 53.7          & 57.9          & 50.0          \\
MaskCLIP \cite{Zhou_2022_MaskCLIP}                 & ViT-L/14                  & 87.3          & 63.2          & 56.3          & 72.1          & 58.6          & 48.9          & 48.5          & 49.3          & 67.4          & 48.2          & 47.6          & 52.4          \\
SCLIP \cite{Wang_2024_SCLIP}                    & ViT-L/14                  & 92.7          & 74.5          & 81.0          & 69.1          & 63.2          & 57.7          & 57.6          & 57.7          & 71.4          & 55.4          & 64.5          & 48.6          \\
ClearCLIP \cite{Lan_2024_ClearCLIP}                & ViT-L/14                  & 92.1          & 74.0          & 80.5          & 68.4          & 63.0          & 57.4          & 52.3          & 63.6          & 70.4          & 54.3          & 61.7          & 48.5          \\
ProxyCLIP \cite{Lan_2014_ProxyCLIP}                & ViT-L/14                  & 94.0          & 75.5          & 86.6          & 67.0          & 64.6          & 57.3          & 52.6          & 62.8          & 73.4          & 57.4          & 65.0          & 51.3          \\
\rowcolor{gray!20} 
\textbf{CLIPer (Ours)}     & \textbf{ViT-L/14}         & \textbf{94.6} & \textbf{86.0} & \textbf{86.7} & \textbf{85.3} & \textbf{68.9} & \textbf{63.3} & \textbf{63.4} & \textbf{64.3} & \textbf{77.9} & \textbf{62.3} & \textbf{69.7} & \textbf{56.3} \\ \bottomrule
\end{tabular}
\caption{Comparison of image-level category classification capability with existing methods. We calculate  classification scores of different categories by max-pooling the segmentation maps, and then calculates the results of mAP, F1, P, and R.  Our method achieves the best results across all datasets, demonstrating its superior performance on category classification.}
\label{tab:pr_ovs}
\vspace{-6pt}
\end{table*}

\subsection{Comparison With Other Methods}

\textbf{On mIoU.} Table  \ref{tab:sota_ovs} compares our proposed method with some state-of-the-art methods on various datasets. 
Our proposed method almost achieves the best performance on all these datasets when using both ViT-B and ViT-L backbones. For instance, on VOC with the ViT-L backbone, SCLIP \cite{Wang_2024_SCLIP} has the mIoU score of 43.5\%, ProxyCLIP \cite{Lan_2014_ProxyCLIP} has the mIoU score of 60.6\%, while our method achieves the mIoU score of 69.8\%. Namely, our method outperforms SCLIP and ProxyCLIP by 26.3\% and 9.2\% on VOC.  On ADE with the ViT-L backbone, ClearCLIP \cite{Lan_2024_ClearCLIP} and ProxyCLIP achieve the mIoU scores of 15.0\% and 22.6\%, while our method achieves the mIoU score of 24.4\%. Namely, our method has the improvements of 9.4\% and 1.8\% on ADE.

\noindent\textbf{On category classification and mask prediction.} Open-vocabulary semantic segmentation can be viewed as two aspects: category classification and mask prediction. To deeply show the advantage of our proposed method on these two aspects, we provide more comparisons with other methods via two experiments. (i) We present image-level precision and recall comparison  to show the advantages of identifying the categories within image. (ii) We compare the segmentation accuracy when giving image-level category labels, demonstrating the advantages of mask prediction. Weakly supervised semantic segmentation aims to train the model based on image-level category labels of training set. By comparing our method with weakly supervised approaches, we can demonstrate the benefits of our method in mask prediction.

In Table \ref{tab:pr_ovs}, we calculate image-level classification scores for all methods, and calculate the results using mAP, F1, P, and R. Our method achieves the best performance on all these metrics. It demonstrate that, our  method can perform better on category classification, which is useful for open-vocabulary semantic segmentation.

\begin{table}[t]
\centering
\footnotesize
\setlength{\tabcolsep}{3.9mm}
\begin{tabular}{l|c|cc}
\toprule
Type                           & Method                & VOC           & COCO          \\ 
\midrule
\multirow{5}{*}{Training-based}      & IRN \cite{Ahn_2019_IRN}                  & 66.5          & 42.4          \\
   & AdvCAM   \cite{Lee_2021_AdvCAM}             & 55.6          & 35.8          \\
   & MCTformer \cite{Xu_2022_MCTFormer}            & 61.7          & -             \\
   & ToCo   \cite{Ru_2023_ToCo}               & 72.2          & -             \\
   & CLIMS   \cite{Xie_2022_Clims}              & 56.6          & -             \\ 
   \midrule
\multirow{5}{*}{Training-free} & CLIP-ES \cite{Lin_2023_CLIPES}     & 70.8    & 39.7   \\
   & DiffSegmenter \cite{Wang_2023_DiffSegmenter}        & 70.5          & -             \\
   & T2M    \cite{Xiao_2023_T2M}               & 72.7          & 43.7          \\
   & iSeg   \cite{Sun_2024_iSeg}               & 75.2          & 45.5          \\
   & \cellcolor{gray!20}\textbf{CLIPer (Ours)} & \cellcolor{gray!20}\textbf{76.9} & \cellcolor{gray!20}\textbf{47.3} \\ 
   \bottomrule
\end{tabular}
\caption{Comparison of pseudo-mask generation with weakly supervised semantic segmentation approaches. Both our method and these weakly supervised approaches predict corresponding pseudo masks according to given image-level category labels. Our CLIPer achieves the best results, showing that our method has better results on mask prediction.}
\vspace{-6pt}
\label{tab:sota_wsss}
\end{table}

Table \ref{tab:sota_wsss} compares our method with some weakly supervised semantic segmentation approaches for pseudo mask generation, where the ground-truth image-level category labels are given. Compared to these weakly-supervised semantic segmentation approaches, our method achieves the best performance. For instance, our method outperforms CLIP-ES \cite{Lin_2023_CLIPES} and iSeg \cite{Sun_2024_iSeg} by 6.1\% and 1.7\%. It demonstrates that, our method can improve mask prediction, and thus improving open-vocabulary semantic segmentation.

\begin{table}[t]
\footnotesize
\centering
\setlength{\tabcolsep}{1.5mm}
\begin{tabular}{l|c|c|cc}
\toprule
Method                                  & Encoder       & Input size            & Time(ms) $\downarrow$ & mIoU $\uparrow$  \\ \midrule
ClearCLIP  \cite{Lan_2024_ClearCLIP}       & ViT-B/16   & $\sim 448\times624$   & 22                    & 51.8  \\
ProxyCLIP \cite{Lan_2014_ProxyCLIP}      & ViT-B/16     & $\sim 336\times468$   & 82                    & 61.3  \\
\rowcolor{gray!20} 
\textbf{CLIPer* (Ours)}                 & ViT-B/16      & $\sim 336\times468$   & \textbf{14}           & 60.1  \\
\rowcolor{gray!20} 
\textbf{CLIPer (Ours)}                  & ViT-B/16      & $\sim 336\times468$   & 158                   & \textbf{65.9}  \\ 
\midrule
ClearCLIP  \cite{Lan_2024_ClearCLIP}     & ViT-L/14     &  $\sim 448\times624$  & 68                    & 46.1  \\
ProxyCLIP \cite{Lan_2014_ProxyCLIP}      & ViT-L/14     & $\sim 336\times468$   & 105                   & 60.6  \\
\rowcolor{gray!20} 
\textbf{CLIPer* (Ours)}                 & ViT-L/14      & $\sim 336\times468$   & \textbf{47}           & 61.2  \\
\rowcolor{gray!20} 
\textbf{CLIPer (Ours)}                  & ViT-L/14      & $\sim 336\times468$   & 192                   & \textbf{69.8}  \\
\bottomrule
\end{tabular}%
\caption{Comparison with other methods in terms of mIoU and inference time on VOC. * denotes the results obtained without  fine-grained compensation. Our CLIPer* has the fastest speed, while Our CLIPer has the best performance.}
\vspace{-6pt}
\label{tab:speed}%
\end{table}%

\begin{figure}[t]
\centering
\includegraphics[width=\linewidth]{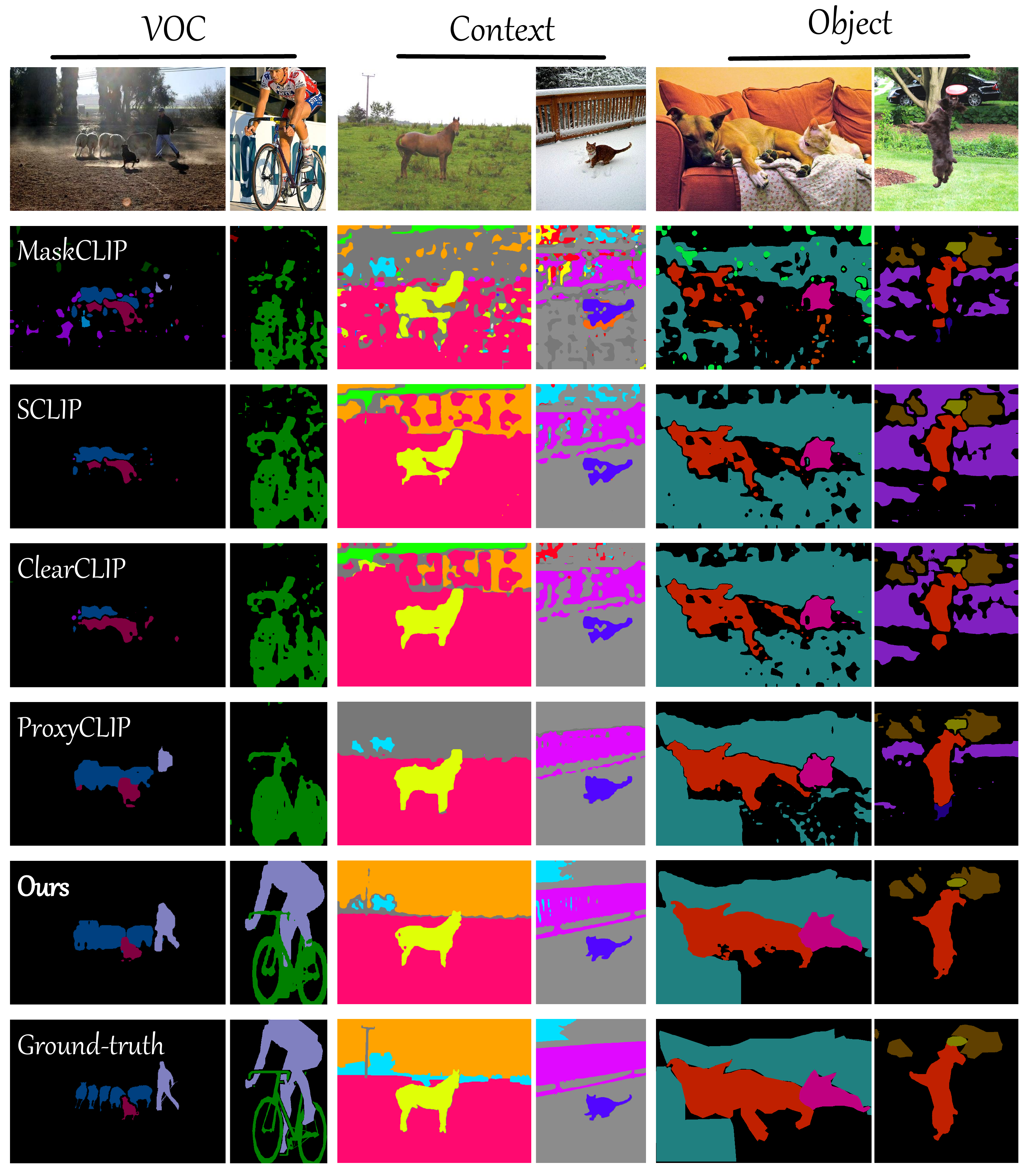}
\caption{Qualitative comparison with existing methods. We show the segmentation results on three different datasets. Compared to these methods, our method has more accurate segmentation results which are closer to the ground-truths.}
\vspace{-6pt}
\label{fig:visual_segmentation}
\end{figure}

\noindent\textbf{Inference time.}  Table \ref{tab:speed} compares inference time and accuracy. Compared to ClearCLIP \cite{Lan_2024_ClearCLIP}, our CLIPer* has faster speed and higher mIoU. Compared to ProxyCLIP \cite{Lan_2014_ProxyCLIP}, our CLIPer* has faster speed and comparable mIoU. Further, our CLIPer with fine-grained compensation significantly improvement the performance of CLIPer*.

\noindent\textbf{Qualitative results.}  Fig. \ref{fig:visual_segmentation} presents some examples of qualitative comparison  on VOC, Context, and Object. Our proposed method has more accurate segmentation maps and precise classification, compared to these methods \cite{maskclip,Lan_2024_ClearCLIP,Lan_2014_ProxyCLIP}. For instance, our method has  finer segmentation on bicycle and correct classification of person in second column, and accurate segmentation on sofa in fifth column. 

\subsection{Ablation study}

\noindent\textbf{Impact of  different modules.} 
Table \ref{tab:ablation} presents the results of integrating difference modules into the baseline. The baseline replaces the original self-attention map at last layer with value-to-value attention map, and removes the feed-forward network (FFN) and residual connections. The baseline achieves the mIoU scores of 51.2\%, 26.5\%, and 32.3\% on VOC, Context, and Object, respectively. When adding early-layer fusion (ELF) module, it has the mIoU scores of 61.2\%, 34.3\%, and 39.6\% on VOC, Context, and Object, outperforming the baseline by 10.0\%, 7.8\%, 7.3\%. When only using fine-grained compensation (FGC) module, it outperforms the baseline by 11.6\%, 3.2\%, 4.1\%.  When integrating the EFL and FGC modules together, it totally has the improvements of 18.6\%, 11.5\%, and 11.0\% on three datasets, respectively. It can significantly demonstrate that, our proposed modules can improve open-vocabulary segmentation performance. 

\begin{table}[t]
\centering
\footnotesize
\setlength{\tabcolsep}{4.6mm}
\begin{tabular}{cc|ccc}
\toprule
ELF                         & FGC                   & VOC           & Context       & Object        \\ 
\midrule
$\usym{2717}$               & $\usym{2717}$          & 51.2          & 26.5          & 32.3          \\
$\usym{2713}$               & $\usym{2717}$          & 61.2          & 34.3          & 39.6          \\
$\usym{2717}$               & $\usym{2713}$          & 62.8          & 29.7          & 36.4     \\
\rowcolor{gray!20} 
$\usym{2713}$               & $\usym{2713}$          & \textbf{69.8} & \textbf{38.0} & \textbf{43.3} \\ \bottomrule
\end{tabular}
\caption{Ablation study of different modules in our CLIPer. ELF represents early-layer fusion, and FGC represents fine-grained compensation. Our proposed modules can significantly improve the performance of the baseline.}
\vspace{-6pt}
\label{tab:ablation}
\end{table}

\begin{table}[t]
\centering
\footnotesize
\setlength{\tabcolsep}{1.0mm}
\begin{tabular}{l|c|ccc}
\toprule
 Attention Type              & Early-layer Embeddings                      & VOC                                   & Context                               & Object                                \\ 
 \midrule
 Query-query       &   $\usym{2717}$     & 50.1               & 22.5             & 28.7       \\
Key-key          & $\usym{2717}$  & 44.8             & 25.4          & 27.0           \\
 Value-value      &  $\usym{2717}$   & 51.2          & 26.5              & 32.3                     \\
Identity matrix    &  $\usym{2717}$   & 40.1           & 22.0             & 25.7                 \\
\rowcolor{gray!20} 
 Early-layer attention    &  $\usym{2717}$  & \textbf{58.6}       & \textbf{33.1}      & \textbf{38.4}  \\ 
\midrule
Query-query        &  $\usym{2713}$      & 55.7          & 26.9          & 32.0          \\
Key-key           &  $\usym{2713}$  & 53.2          & 29.5          & 33.1          \\
Value-value       &   $\usym{2713}$     & 54.8          & 29.4          & 34.4          \\
Identity matrix   &   $\usym{2713}$    & 43.3          & 24.8          & 27.8          \\
\rowcolor{gray!20} 
Early-layer attention  & $\usym{2713}$  & \textbf{61.2} & \textbf{34.3} & \textbf{39.6} \\ 
\bottomrule
\end{tabular}
\caption{Impact of different designs in our early-layer fusion. Our early-layer fusion contains fusing the embeddings and attention maps. In the top part, we  compare our fused attention with some self-self attention operations. In the bottom part, we feed the early-layer embeddings to the attention map at last layer.}
\vspace{-6pt}
\label{tab:abltation_esf}
\end{table}

\begin{table}[t]
\centering
\footnotesize
\setlength{\tabcolsep}{5.5mm}
\begin{tabular}{l|ccc}
\toprule
Type        & VOC                      & Context       & Object        \\ 
\midrule
Single      & 65.3                     & 36.1          & 41.2          \\
Mean        & 64.9                     & 36.1          & 41.0          \\
\rowcolor{gray!20} 
Multiplication
        & \textbf{69.8} 
        & \textbf{38.0} & \textbf{43.3} \\ \bottomrule
\end{tabular}
\caption{Comparison of different strategies using the multi-head attention maps in Stable Diffusion. Single represents that, we evaluate each single head  and report the best-performing head. Mean represents that, we average multi-head attention maps, and multiplication presents that, we perform matrix multiplication to fuse multi-head attention maps.}
\vspace{-6pt}
\label{tab:abltation_refinement}
\end{table}

\noindent\textbf{Effect of early-layer fusion.} Table \ref{tab:abltation_esf} compares our early-layer fusion with some self-self attention operations. Our early-layer fusion module fuses both the patch embeddings of early layers and the attention maps of early layers. In the top part, we compare our early-layer fused attention with these self-self attention operations, which all take the embeddings at last layer as input. Compared to these self-self attention operations, our early-layer fused attention has the best performance. For instance, on VOC, Context, and Object, the query-query attention achieves the mIoU scores of 50.1\%, 22.5\%, and 28.7\%, while our early-layer fused attention has the  mIoU scores of 58.6\%, 33.1\%, and 38.4\%. Namely, our early-layer fused attention outperforms query-query attention by 8.5\%, 10.6\%, and 9.7\%, respectively.

In  bottom part, we show the impact of feeding early-layer embeddings to last layer. When integrating early-layer embeddings into early-layer attention, it has the improvements of 2.6\%, 1.2\%, and 1.2\% on VOC, Context, and Object, respectively. We observe that, using early-layer embeddings can also improve the performance of existing self-self attention operations. For instance, when combing it with value-value attention, it has the improvements of 3.6\%, 2.9\%, and 1.1\% on VOC, Context, and Object, respectively.

\begin{figure}[t]
\centering
\includegraphics[width=\linewidth]{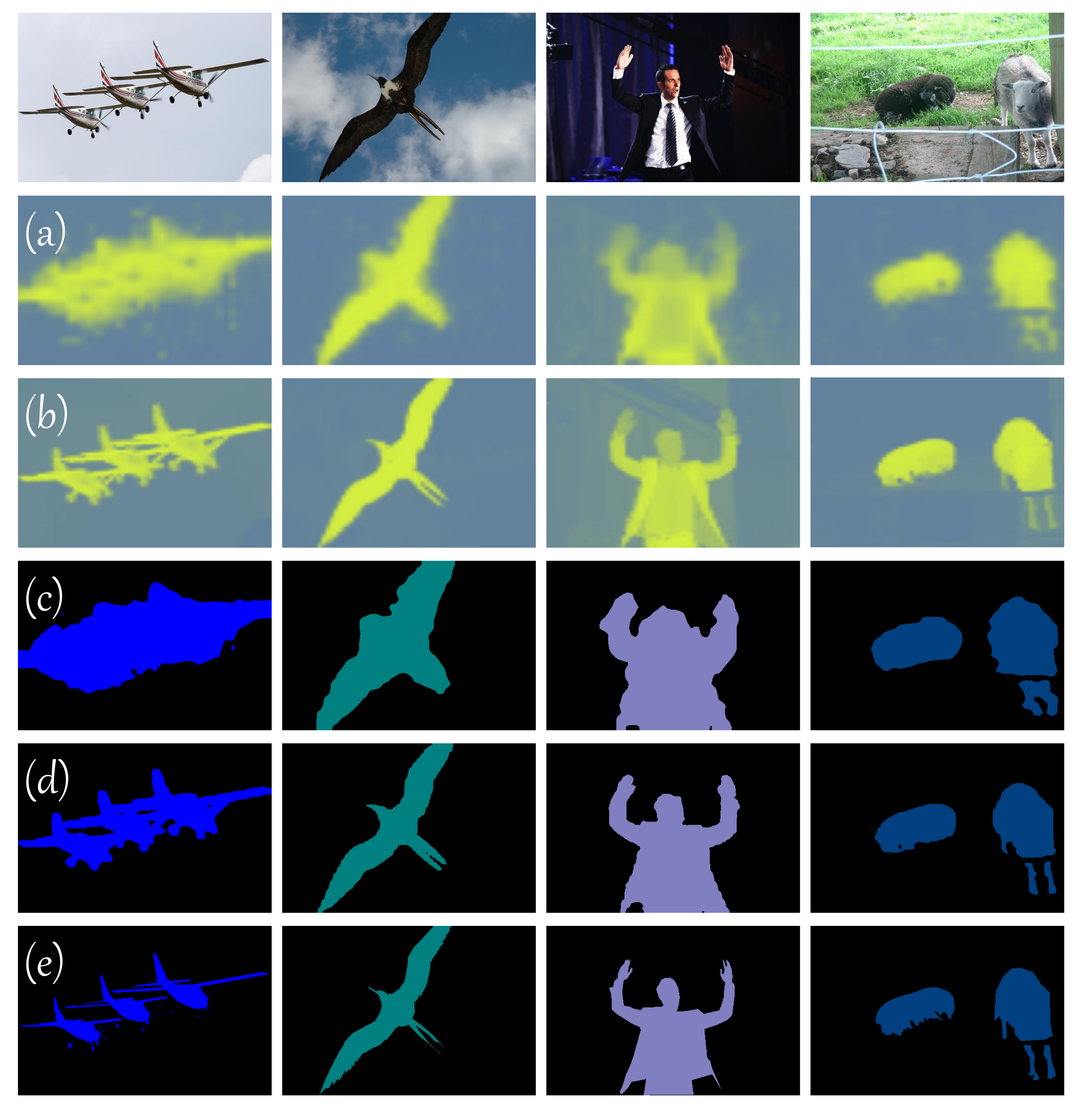}
\caption{Visualization of fine-grained compensation. Given the input images in the top, we present the attention maps before and after fine-grained compensation in (a) and (b), show the binary segmentation maps in (c) and (d). We also present the ground-truth segmentation maps in (e).}
\vspace{-6pt}
\label{fig:visual_refine}
\end{figure}

\noindent\textbf{Effect of fine-grained compensation.} Table \ref{tab:abltation_refinement} presents different strategies to fuse the multi-head attention maps in Stable Diffusion for fine-grained compensation, including  a single head (Single), averaging all heads (Mean), and combining all heads with matrix multiplication (multiplication). Compared to the baseline, all  three strategies can improve the performance, demonstrating that the attention maps in Stable Diffusion can improve the CLIP-based segmentation. Among these  strategies, matrix multiplication has the best performance, which is adopted as our final setting.

Fig. \ref{fig:visual_refine} shows some visualized examples  before and after fine-grained compensation. Before  fine-grained compensation, the attention maps in (b)  provide coarse spatial structure information of objects. By using our fine-grained compensation, the attention maps in (c) are able to provide more accurate responses around  object contour. As a result, compared to that in (d), using our fine-grained compensation has more accurate segmentation maps in (e). It  demonstrates that, our fine-grained compensation  can improve local details of corse segmentation maps by early-layer fusion.

\section{Conclusion}

This paper presents CLIPer, a novel training-free method  to hierarchically improve the spatial representation of CLIP for open-vocabulary semantic segmentation. To achieves this goal, we design two components, including early-layer fusion and fine-grained compensation. The early-layer fusion aims to improve spatial coherence of output patch embeddings by using the early-layer information of patch embeddings and attention maps. The fine-grained compensation module employs the fine attention maps of diffusion model to futher improve local details of segmentation maps. Our proposed method achieves the superior performance on various public segmentation datasets. We still observe that, our proposed method struggles from accurately segmenting the tiny objects. In the future, we will explore how to adapt the pre-trained model with high-resolution input for improving tiny object segmentation.

{
    \small
    \bibliographystyle{ieeenat_fullname}
    \bibliography{main}
}


\end{document}